\documentclass{article}
\usepackage{spconf,amsmath,graphicx}
\usepackage{amssymb}
\usepackage{booktabs}
\usepackage{multirow}
\usepackage{xcolor}
\usepackage[ruled, lined,noend, linesnumbered, commentsnumbered]{algorithm2e}
\usepackage{times}
\usepackage{epsfig}

\graphicspath{{Figs/}}

\usepackage[pagebackref=true,breaklinks=true,letterpaper=true,colorlinks,bookmarks=false]{hyperref}

\newcommand{\mysection}[1]{\vspace{2pt}\noindent\textbf{#1}}

\definecolor{commentsColor}{RGB}{219, 48, 122}

\SetCommentSty{mycommfont}
\let\oldnl\nl
\newcommand{\nonl}{\renewcommand{\nl}{\let\nl\oldnl}}

\usepackage[normalem]{ulem}
\useunder{\uline}{\ul}{}
\makeatletter
\DeclareRobustCommand\onedot{\futurelet\@let@token\@onedot}
\def\@onedot{\ifx\@let@token.\else.\null\fi\xspace}

\def\eg{\emph{e.g}\onedot}

\def\etal{\emph{et al}\onedot}
\makeatother
\title{Efficient Semantic Segmentation for Aerial Imagery Using Query Points and Superpixel Supervision}
\name{Santiago Rivier \qquad Carlos Hinojosa \qquad Silvio Giancola \qquad Bernard Ghanem}
\address{King Abdullah University of Science and Technology (KAUST)\\
		{\tt\footnotesize santiago.rivier@gmail.com, \{carlos.hinojosa, silvio.giancola, bernard.ghanem\}@kaust.edu.sa}}

\begin{document}
%
\maketitle
%

\begin{abstract}
Semantic segmentation is crucial in remote sensing, where high-resolution satellite images are segmented into meaningful regions. Recent advancements in deep learning have significantly improved satellite image segmentation. 
However, most of these methods are typically trained in fully supervised settings that require high-quality pixel-level annotations, which are expensive and time-consuming to obtain. 
In this work, we present a weakly supervised learning algorithm to train semantic segmentation algorithms that only rely on query point annotations instead of full mask labels. Our proposed approach performs accurate semantic segmentation and improves efficiency by significantly reducing the cost and time required for manual annotation.
Specifically, we generate superpixels and extend the query point labels into those superpixels that group similar meaningful semantics. Then, we train semantic segmentation models supervised with images partially labeled with the superpixel pseudo-labels.
We benchmark our weakly supervised training approach on an aerial image dataset and different semantic segmentation architectures, showing that we can reach competitive performance compared to fully supervised training while reducing the annotation effort. The code of our proposed approach is publicly available at: \url{https://github.com/santiago2205/LSSQPS}.
\end{abstract}

\begin{keywords}
Weakly supervised learning, Aerial Imagery, Semantic Segmentation, Remote Sensing Images.
\end{keywords}

\section{Introduction}
\label{sec:intro}



Semantic segmentation is a critical task in remote sensing and computer vision, where the goal is to assign a class label to each pixel in an image. In particular, semantic segmentation of satellite and aerial images could be useful for urban planning, disaster recovery, autonomous agriculture, environmental monitoring, and many other applications. Recent supervised deep-learning approaches have achieved remarkable performance in the semantic segmentation of satellite and aerial images. Some recent works for semantic segmentation include SegNet, FCN, U-NET, and PSPNet with excellent results in satellite imagery~\cite{audebert2017joint,marcos2018land,sherrah2016fully}. 

Typically, these approaches require extensive pixel-level annotations for supervision, which is expensive and time-consuming. To address the challenge of limited labeled data, a new learning paradigm relying on weaker types of labels, namely Weakly Supervised Learning (WSL), has been proposed. These methods leverage weaker forms of annotations, such as bounding-box annotations, scribbles, points, or 
simple image-level class labels to infer the segmented regions.


In general, these approaches can be especially promising in the field of remote sensing, 
where acquiring extensive volumes of satellite and aerial images is common, yet densely annotating them at a pixel level is impractical. 
However, training models with very few supervised pixels poses a significant challenge in correctly identifying the spatial extent of objects in the image. Specifically, the model could only focus on a small section of the target object, resulting in incomplete or inaccurate segmentation. Furthermore, it could also predict all pixels as part of the background class, leading to a false negative prediction~\cite{bearman2016s}.


This paper presents an efficient semantic segmentation approach for aerial imagery using just a few pixel-level annotations. To overcome the limitations of point-based WSL methods, we rely on superpixels to provide better supervision by embedding consistent semantic information and spatial context into our model. Our method yields comparable results to fully supervised approaches, reducing annotation effort while maintaining high segmentation accuracy. This achieves a better trade-off between annotation efficiency and segmentation performance.

We summarize our contributions as follows:


\begin{enumerate}
    \item We present a novel methodology for weakly annotated settings that extend labels from points to superpixels and trains on partially annotated superpixels.
    \item We present a novel weighted masked loss function, which only computes labeled pixel and weight balance of each class.
    \item We provide a comprehensive analysis of our WSL approach, showing its capability on different datasets and different model architectures.
\end{enumerate}

\vfill

%




\section{Related Work}
\label{sec:related_works}



\mysection{Fully-Supervised Semantic Segmentation.} Traditional methods extract spectral and spatial information via a preprocessing step and rely on traditional supervised algorithms like SVM, MRF, and KNN~\cite{huang2016spectral}. In recent years, several deep neural networks have been developed to extract high-level features of satellite images, achieving state-of-the-art supervised semantic segmentation performance~\cite{9756442}. 
Some early methods include U-Net~\cite{ ulmas2020segmentation} and DeepLabV3~\cite{chen2017rethinking}, which can be adapted to suit the unique characteristics of satellite images, with their high-resolution and multi-spectral nature. Among recent works, Miao~\etal~\cite{miao2023mafnet} propose a multiangle attention fusion network for the classification of land cover types in aerial images. The proposed network in \cite{miao2023mafnet} uses a 50-layer residual network as a feature extraction network and adds an adaptive special-shaped window attention module to the deep layer of the network to extract deep semantic information, building the connection between global information.
In general, the success of recent deep learning approaches hinges on a large amount of labeled datasets~\cite{demir2018deepglobe,8900532,boguszewski2021landcover, tao2022msnet}. While satellite images are readily obtainable, the efforts to generate high-quality datasets are limited by the enormous effort required to create accompanying annotations, which are not always available and often prohibitively expensive to acquire. To address these challenges, researchers have proposed alternative approaches, such as unsupervised and weakly-supervised semantic segmentation, which may require fewer labeled examples and potentially lead to models with better generalization capabilities.

\begin{figure*}[t]
    \centering
    \includegraphics[width=\linewidth]{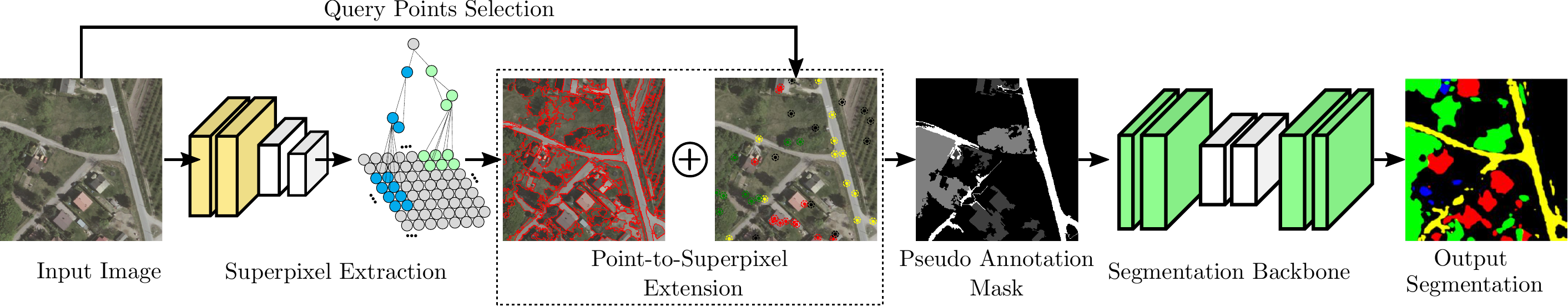}
    \caption{
    \textbf{Proposed WSL Approach for Semantic Segmentation on Aerial Images.}
    Our approach solely considers query point-based annotation on satellite images and relies on superpixel extraction to extend the point-based annotation into larger regions. We minimize our proposed masked loss by leveraging the generated partial mask pseudo-annotations, which provide more supervisory signals than the sole query point-based annotations.}
    \label{fig:pipeline_figure}
    \vspace{-7pt}
\end{figure*}

\mysection{Unsupervised Semantic Segmentation}. 
Unsupervised learning involves identifying and labeling different objects and regions in satellite images \textit{without} relying on pre-existing labeled datasets. Recent advances in unsupervised satellite image segmentation include fast subspace clustering~\cite{hinojosa2018coded, hinojosa2021hyperspectral, hinojosa2021fast, lopez2021efficient} and deep clustering approaches~\cite{caron2018deep, caron2020unsupervised, saha2022unsupervised}. Although unsupervised semantic segmentation algorithms do not require any labeled data, they are generally less accurate than supervised approaches as they may not capture the high-level semantic meaning of the image. 


\mysection{Weakly Supervised Semantic Segmentation.} 
Unlike traditional supervised learning methods that require pixel-level annotations for every image, weakly supervised segmentation relies on less precise annotations such as point-level labels~\cite{bearman2016s, wang2020weakly, moliner2020weakly}, image-level labels~\cite{ahn2018learning, wang2020weakly}, scribbles~\cite{ferreira2022weaklier} or bounding boxes~\cite{khoreva2017simple}. For instance, Ferreira ~\etal~\cite{ferreira2022weaklier} propose a pipeline to automatically generate scribbles in images, requiring that the user only label 10\% of the images in a given dataset and a semantic segmentation pipeline that uses only images with scribbles to train a network. Wang~\etal~\cite{wang2020weakly} explores weak labels in the form of a single-pixel label per image and class activation maps to perform semantic segmentation on satellite images. 
On the other hand, authors in \cite{hamilton2022unsupervised} recently propose STEGO (Self-supervised Transformer with Energy-based Graph Optimization), which is an unsupervised approach that leverages the intrinsic structure within the data to learn features and then perform cluster compactification. In general, one issue with training models with very few or no supervised pixels is correctly inferring the spatial context of the objects. Indeed, weakly supervised methods are prone to local minima: focusing on only a small part of the target object or predicting all pixels as belonging to the background class~\cite{bearman2016s}. 
In this work, we leverage point-based annotations along with superpixels to train semantic segmentation algorithms. Specifically, our proposed approach extends point annotations into superpixels to create a partial mask for supervision. We refine our end-to-end framework by leveraging a pretrained segmentation backbone, which we further fine-tune using our proposed masked loss to particularly learn from the points selected by the user.

\vspace{-7pt}
\section{Methodology}
\label{sec:prop_method}

We present a novel end-to-end framework for satellite image semantic segmentation. Our proposed method, shown in Fig.~\ref{fig:pipeline_figure}, utilizes user-selected query points $P$ as the sole source of supervision for the semantic segmentation task. The framework consists of three stages: superpixel extraction, pseudo-mask generation by expanding point annotations into superpixels, and the segmentation module, which is fine-tuned with our proposed masked loss. 

\vspace{-5pt}
\subsection{Point selection}

The input of our semantic segmentation algorithm is the image and the corresponding set of $P$ query points as input. 
%
To ease the experimentation and validation of our method, we employ two methods for selecting the points: random selection per image and per class label. In the first approach, we randomly select pixels as query points and assign the corresponding class label. However, this approach may lead to the under-sampling of pixels from specific classes, especially in imbalanced datasets. In contrast, the second approach evaluates the number of classes in each image and distributes the required points equally among the classes. This approach ensures accurate annotation for imbalanced classes.

\vspace{-5pt}
\subsection{Superpixel}

In this work, we adopt the DAL-HERS~\cite{peng2022hers} superpixels algorithm to generate the superpixel regions shown in Fig.~\ref{fig:pipeline_figure}. DAL-HERS is a deep learning-based technique that computes affinities between neighboring pixels. Specifically, the algorithm is based on a hierarchical merging approach, where neighboring regions are merged into larger superpixels based on their affinities.
The affinities between neighboring pixels are computed using a deep neural network trained to predict the probability of pixel pairs belonging to the same superpixel. The network considers the image's color and texture information to compute the affinities.
The merging process is guided by the hierarchical entropy rate \cite{peng2022hers}, which measures the complexity of the image at different scales. The algorithm seeks to minimize the hierarchical entropy rate by merging neighboring superpixels with high affinities.



\vspace{-5pt}
\subsection{Weakly Supervised Training from Superpixel}
To generate the pseudo-annotations mask, we need to merge the query points selected by the user with the superpixel regions. Here, the objective is to extend the point-level annotations to regions considering the spatial contextual information. The underlying logic of our approach is that if we have a superpixel region with one point labeled, we propagate this label in all this region. However, the same superpixel region could contain more than one point, \eg, if the superpixel algorithm makes any mistake in detecting the object boundaries. 
Therefore, it is possible to have more than two labels in the same superpixel region when looking at the ground truth. To solve this, we count the number of points by each class and propagate the class with the most points. For the rest of the regions that do not contain any query points, we assign such regions' pixels to the background class.


\vspace{-5pt}
\subsection{Weighted Masked Loss}

We propose a novel weighted masked loss to train our proposed approach and leverage the supervisory signals provided by the generated pseudo-annotations mask. This loss function consists of a Mean Square Error (MSE), which we multiply with a binary mask that contains one where the image is labeled and zero where the image is unlabeled. With this approach, we consider only the pixels with labels during our optimization process. Furthermore, this loss function addresses class imbalance by penalizing classes that are overrepresented, effectively balancing the influence of each class. Formally, we define our proposed weighted masked loss as:
\begin{equation*}
\mathcal{L} = \frac{1}{N} \sum{l_{i}=} \frac{1}{N} \sum_{i=1}^{N} \left( \frac{1}{C} \sum_{c=1}^{C} \left[ m_{i,c} \cdot (wl_{i,c} - pl_{i,c})^2 \right] \right),
\end{equation*}
\noindent where $N$ corresponds to the batch size; $l_i$ is the loss by sample, $C$ is the number of classes; $m_{i,c}$ is the binary mask value for class $c$ and image sample $i$; $wl_{i,c}$ is the weakly supervised label; and $pl_{i,c}$ is the predicted label. Note that the variable $pl_{i,c}$ represents the estimated label and clarifies the use of one-hot encoding for class representation.
During training, we use our loss $\mathcal{L}$ with different semantic segmentation models and feature extractor backbones; see Section \ref{sec:ablation}.
\vspace{-7pt}
\section{Experiments}
\label{sec:results}

\subsection{Experimental Setup}

\mysection{Dataset.} We evaluated our proposed weakly supervised semantic segmentation approach using the LandCoverAI dataset \cite{boguszewski2021landcover}. The images were captured from 2015 to 2018 with a spatial resolution of $25$ or $50$ cm per pixel and contain three spectral bands (RGB). The dataset consists of $41$ orthophoto tiles manually selected from counties across Poland, with each tile covering approximately $5 \ \text{km}^2$ area. The LandCoverAI dataset includes four classes: \textbf{\textit{\textcolor{red}{Buildings}}} (1), \textbf{\textit{\textcolor{green}{Woodland}}} (2), \textbf{\textit{\textcolor{blue}{Water}}} (3), and \textbf{\textit{\textcolor{yellow}{Road}}} (4). A \textit{building} annotation includes the roof and visible walls, while \textit{woodland} denotes land covered with multiple trees. \textit{Water} encompasses both flowing and stagnant water bodies, excluding ditches and dry riverbeds. The \textit{road} annotation includes roads, parking areas, unpaved roads, and tracks. Lastly, the \textbf{\textit{\textcolor{black}{Background}}} class represents areas not classified into any of the above classes, which may include fields, grass, pavements, etc.
In this work, we use TorchGeo \cite{Stewart_TorchGeo_Deep_Learning_2022} to load and preprocess the LandCoverAI images. TorchGeo is a PyTorch library that provides datasets and pre-trained models specific to geospatial data. Specifically, TorchGeo provided us with $10674$ images of size $512\times 512$, which we split into $60$\% for training, $20$\% for validation, and $20$\% for testing.




\mysection{Models.}
In principle, any semantic segmentation backbone can be used within our framework, as illustrated in Figure \ref{fig:pipeline_figure}. In our experiments, we tested three main architectures.

\underline{FCNet \cite{long2015fully}} stands for \textit{Fully Convolutional Network} and is a well-known architecture to perform semantic segmentation. The main advantage of FCNet is that it can accept inputs of arbitrary size and produce correspondingly-sized output with efficient inference and
learning. The architecture of FCNet employs solely locally connected layers, such as convolution, pooling, and upsampling. Avoiding dense layers means fewer parameters, making the networks faster to train. The network consists of a downsampling path, used to extract and interpret the context, and an upsampling path, which allows for localization. FCNs also use skip connections to recover the fine-grained spatial information lost in the downsampling path.


\underline{UNet} \cite{ronneberger2015u} is a neural network named after its U-shape design, comprising a contracting path and an expansive path. The contracting path, similar to a typical convolutional neural network, involves convolutional and pooling layers to gradually reduce the spatial dimensions and capture global context and high-level features. Conversely, the expansive path employs upsampling and convolutional layers to progressively increase the feature map's spatial dimensions, integrating it with the feature maps from the contracting path. This process refines the segmentation mask by combining local information with a global context.
An essential aspect of UNet is its skip connections, facilitating the fusion of feature maps from the contracting and expansive paths. This enables the network to capture both local and global information, contributing to accurate segmentation. These skip connections help address the problem of information loss during downsampling.

\underline{DeepLabV3} \cite{chen2017rethinking} is a family of convolutional models designed for semantic segmentation tasks. DeepLabV3 uses a modified version of the atrous convolution, also known as dilated convolution, to extract features at multiple scales from the input image.
The DeepLabV3 architecture incorporates a backbone to extract complex features from the input image. In this work, we explore the combination of DeepLabV3 with ResNet50 and ResNet101 networks as backbones to see its impact on our weakly supervised semantic segmentation approach. The extracted features from the input image are then passed through a series of convolutional and pooling layers to reduce the spatial resolution of the feature maps. The resulting feature maps are then fed into the atrous convolutional module of DeepLabV3 to extract features at multiple scales.
Finally, the resulting features are fed into a classifier that produces a segmentation mask, where each pixel is assigned a label indicating its corresponding class.

\mysection{Metrics.}
To measure the performance of our proposed framework, this work uses the mean Intersection-Over-Union (mIoU) metric. Here, we used the Jaccard index from the PyTorch library. The Jaccard index is a statistic that can be used to determine the similarity and diversity of a sample set. It is defined as the size of the intersection divided by the union of the sample sets:
{
\setlength{\belowdisplayskip}{4pt}
\setlength{\abovedisplayskip}{4pt}
\begin{equation*}
    J(A,B) = {|A \cap B|}/{|A \cup B|}
\end{equation*}
}

To configure the parameters of the Jaccard index, we ignore index zero (background class) as our loss function is only computed from the labeled regions. Also, the statistics used to calculate the mIoU are set as ``micro'', which means that the sum statistics are computed over all labels.

\mysection{Training details.}
We implemented our end-to-end framework in Pytorch. We trained models from scratch during $300$ epochs and fine-tuned pre-trained models during $60$ epochs with a batch size of $32$ using the Adam optimizer. The learning rate (LR) was set to $1e^{-4}$ with a scheduler that decreased the LR by a factor of 10 at each plateau until convergence. All experiments were conducted on two NVIDIA V100 GPUs.

\vspace{-5pt}
\subsection{Ablation Study}
\label{sec:ablation}
We employed the DeepLabV3 architecture as the segmentation backbone to conduct the ablation study of our proposed method. We repeated each experiment five times to ensure reliable and consistent results.

\begin{table}[t!]
    \centering
    \caption{
    \textbf{Multiple ablation studies of our method}.
    (a) using the DeepLabV3 backbone model with cross-entropy loss and our proposed masked loss; 
    (b) varying the number of points to generate the pseudo annotation mask; 
    (c) varying the number of superpixels after fixing the number of points to 50; and
    (d) with/without the edge option in DAL-HERS.}
    \vspace{2pt}
    \label{table:1}
    \def\arraystretch{1}%
    \begin{tabular}{@{}l|c|c@{}}
    \toprule
    \multicolumn{2}{c|}{\bf Experiment}& \bf mIoU         \\ \midrule
    \multicolumn{1}{l|}{\parbox[t]{10mm}{\multirow{3}{*}{\rotatebox[origin=c]{90}{\begin{tabular}[c]{@{}c@{}}Loss\\(a)\end{tabular}}}}}          & Cross Entropy       & $0.5646 \pm 0.02$  \\
      & Masked Loss (Ours) & $\mathbf{0.7564 \pm 0.01}$  \\ 
    & Fully Labeled  & $0.8399 \pm 0.01$ \\ 
      \midrule
    \multicolumn{1}{l|}{\parbox[t]{3mm}{\multirow{5}{*}{\rotatebox[origin=c]{90}{\begin{tabular}[c]{@{}c@{}} Point\\(b)\end{tabular}}}}} 
    & $10$           & $0.6690 \pm 0.03$ \\
    & $20$           & $0.7194 \pm 0.01$ \\
    & $30$           & $0.7433 \pm 0.01$ \\
    & $40$           & $0.7384 \pm 0.02$ \\
    & $50$           & $\mathbf{0.7564 \pm 0.01}$ \\
    \midrule
    \multicolumn{1}{l|}{\parbox[t]{3mm}{\multirow{4}{*}{\rotatebox[origin=c]{90}{\begin{tabular}[c]{@{}c@{}} Superpixel\\(c)\end{tabular}}}}}                                    & $80$                                & $0.7543 \pm 0.01$ \\
      & $100$                               & $\mathbf{0.7564 \pm 0.01}$ \\
      & $200$                               & $0.6983 \pm 0.03$ \\
      & $300$                               & $0.6967 \pm 0.05$ \\ \midrule
    \multicolumn{1}{l|}{\parbox[t]{3mm}{\multirow{2}{*}{\rotatebox[origin=c]{90}{\begin{tabular}[c]{@{}c@{}} Edge
    \\ (d)\end{tabular}}}}}                                                      & True                                & $0.7527   \pm 0.01$ \\
      & False                               & $\mathbf{0.7564 \pm 0.01}$ \\ \bottomrule
    \end{tabular}
    \vspace{-5pt}
\end{table}

\begin{figure}[t!]
    \centering
    \includegraphics[width=\columnwidth]{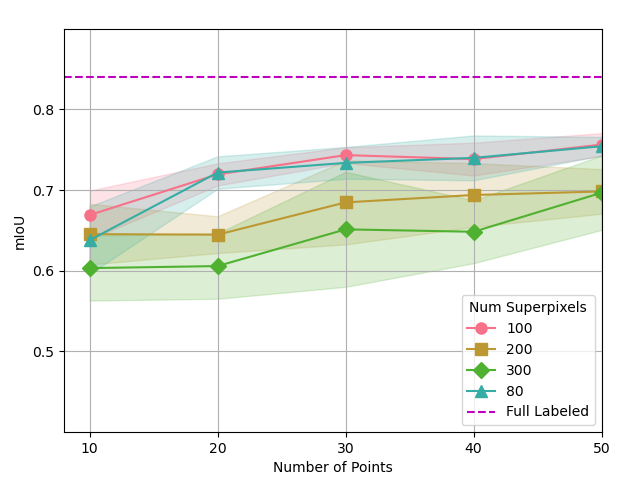}
    \caption{\textbf{Segmentation Performances per Number of Points.}
    We show the performance with different numbers of points for each number of superpixels.}
    \label{fig:iou_vs_npoints}
    \vspace{-7pt}
\end{figure}

\mysection{Effect of the Loss.}
Developing a robust loss function is essential for achieving accurate image segmentation. In our approach, we tested PyTorch's MSELoss and CrossEntropyLoss with various hyperparameters, but they yielded unsatisfactory results. Therefore, we introduced our weighted masked loss (Section \ref{sec:prop_method}) that takes into account the prediction, one-hot encoded label, and label without encoding. By computing the MSE loss only for labeled pixels, we observed significant improvements in our experiments. To understand the impact on uncomputed regions, which may hold crucial information, we compared model performance using completely labeled and weakly labeled data, as depicted in Fig. \ref{fig:iou_vs_npoints}. This analysis shed light on the loss function's influence on uncomputed regions and overall performance. To address class imbalance, we implemented class balancing by dividing the loss result by the percentage of each class. This approach penalizes high-occurrence classes while rewarding low-occurrence ones, leading to enhanced image segmentation accuracy.


\mysection{Effect of the number of point labels.}
Using more query points in our framework will conduct in a better pseudo annotation mask which means more level of supervision. Therefore, as observed in Table \ref{table:1}, increasing the number of points will lead to better performance in terms of mIoU.

\mysection{Effect of the Number of Superpixels.}
The number of superpixels employed in our framework significantly influences its performance. A reduction in the number of superpixels results in larger regions while increasing the number of superpixels leads to smaller regions. The operation of DAL-HERS imposes a minimum number of superpixels required for successful image segmentation. This minimum threshold is determined empirically and depends on factors such as image size and complexity. For the LandCover dataset used in our study, we found that a minimum of 80 superpixels is necessary.
Furthermore, our experimental results, as depicted in Fig. \ref{fig:iou_vs_npoints}, reveal an inverse relationship between the number of superpixels and mIoU. Specifically, an increase in the number of superpixels corresponds to a decrease in mIoU.

\mysection{Effect of using edge in superpixel.}
The current superpixel algorithm provides the option of specifying the ``edge'' hyperparameter to use a border detector and generate more precise superpixel regions. However, the default setting for this parameter is False. In our experiments, we compared the performance of superpixels generated with and without the edge parameter on a dataset of interest. Specifically, we compared the mIoU scores of $100$ superpixels with 50 points each in Table \ref{table:1}(d). As observed from the table, there were no significant performance improvements when using the edge parameter.

\begin{figure}[t]
    \centering
    \includegraphics[width=\columnwidth]{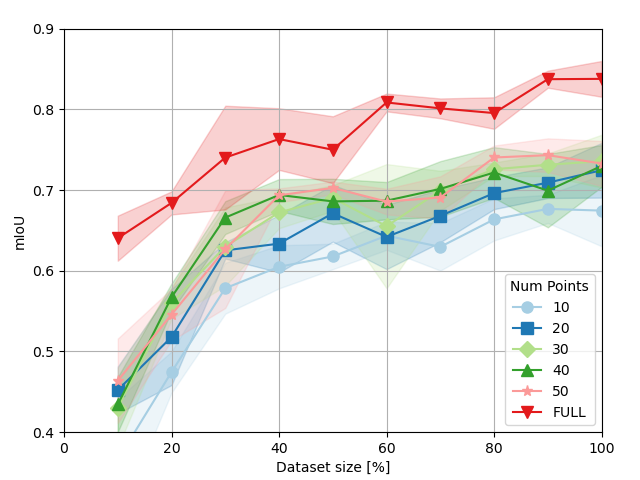}
    \caption{
    \textbf{Segmentation Performances per Dataset Size.}
     Our results demonstrate that larger datasets lead to improved performance. The experiment was conducted using a dataset of a total amount of $10,674$ images.
    }
    \label{fig:miou_vs_dataset_size}
    \vspace{-7pt}
\end{figure}

\mysection{Model Performance Varying Dataset Sizes.} We analyze how the performance of our method changes when varying the dataset size in terms of mIoU. Fig. \ref{fig:miou_vs_dataset_size} shows the relationship between mIoU and the dataset size used in our method.

\mysection{Generalization to different segmentation backbones}. To evaluate the generalizability of our weakly supervised semantic segmentation approach, we conducted experiments using three widely used semantic segmentation models as backbones: DeepLabV3, U-Net, and FCNet. To better assess the performance of DeepLabV3 and FCNet models, we compare them using different encoders: ResNet50 and ResNet101. The results shown in Table \ref{tab:segmentation_backbones} show that the semantic segmentation model DeepLabV3 obtained the best performance regardless of the two variants of encoders.

\begin{figure}[t!]
    \centering
    \includegraphics[width=\columnwidth]{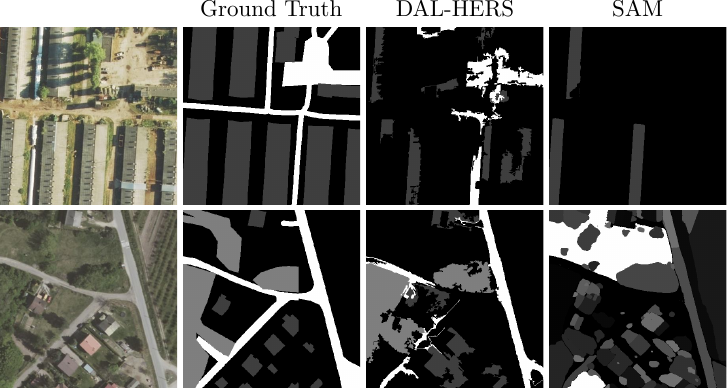}
    \caption{\textbf{DAL-HERS vs SAM.} 
    Comparison of the DAL-HERS and SAM model outputs after labeling the images with points.}
    \label{fig:dal-hers_sam}
    \vspace{-15pt}
\end{figure}

\mysection{Superpixel extraction model}. Although superpixel models are not the primary focus of our work, we also performed a comparative analysis of the \textit{pre-trained} models DAL-HERS and SAM using the same satellite image to assess their segmentation performance. Our findings reveal that DAL-HERS outperforms SAM in segmentation accuracy. Notably, as seen in Fig \ref{fig:dal-hers_sam},  DAL-HERS demonstrates the ability to avoid segmenting undesired areas, which SAM fails to achieve. More importantly, DAL-HERS provides an accurate segmentation of the boundaries with up to three times faster processing time than SAM. This is a crucial feature of our method, especially when dealing with extensive image datasets.

\begin{table}[t!]
    \caption{
    \textbf{Backbone generalization.}
    Our WSL approach generalizes well across different segmentation backbones (see Fig. \ref{fig:pipeline_figure}), with DeepLabV3 performing the best.
    }
    \vspace{2pt}
\centering
\def\arraystretch{1}%
\begin{tabular}{@{}l|c@{}}
\toprule
\textbf{Model} & \textbf{mIoU}   \\\midrule
DeepLabV3 (ResNet50)    & $\mathbf{0.756 \pm 0.01}$\\
DeepLabV3 (ResNet101)   & $\mathbf{0.756 \pm 0.02}$\\
FCN (ResNet50)          &$0.728 \pm 0.01$ \\
FCN (ResNet101)         &$0.735 \pm 0.01$ \\
U-Net                   & $0.589 \pm 0.02$\\ 
\bottomrule
\end{tabular}
\label{tab:segmentation_backbones}
\vspace{-15pt}
\end{table}

\begin{table}[t!]
    \centering
    \caption{Comparison with other methods. Values correspond to mIoU in percentage.}
    \vspace{2pt}
    \resizebox{\columnwidth}{!}{%
        \begin{tabular}{cccc|ccc}
    \toprule
    \multicolumn{4}{c|}{Fully Supervised}                         & \multicolumn{3}{c}{Weakly Supervised}                                       \\ \midrule
    SGBNet & DBPNet & MAFNet                           & Ours FSL & Scribble & STEGO & Ours                            \\
    85.479 & 86.00  & \textbf{87.133} & 83.99    & 49.90        & 70.24 & \textbf{75.64} \\ \bottomrule
    \end{tabular}}
    \label{tab:comparison}
    \vspace{-15pt}
\end{table}

\begin{figure}[t]
    \centering
    \includegraphics[width=\columnwidth]{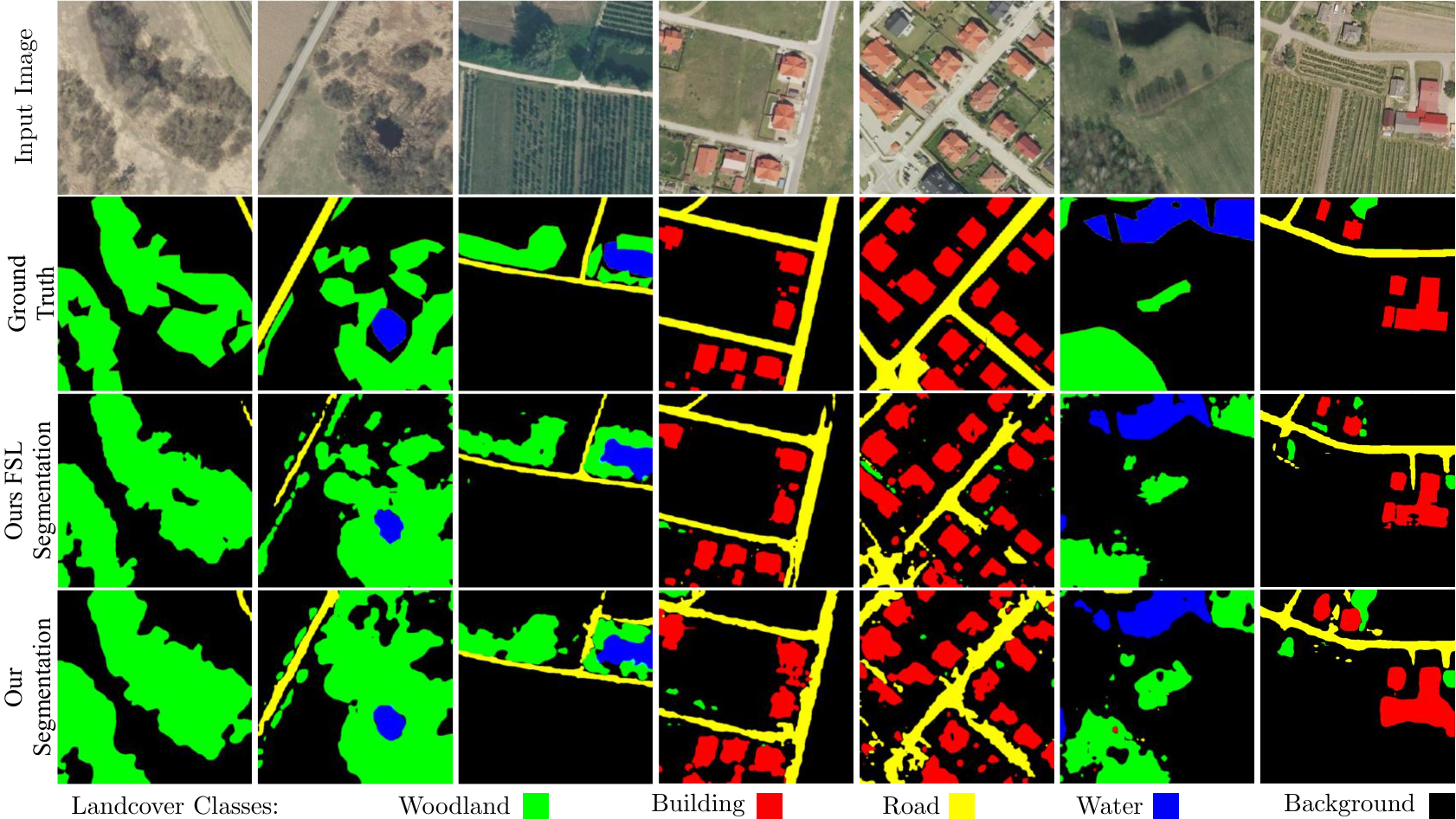}
    \caption{
    \textbf{Qualitative Results on LandCoverAI.} 
    Top to bottom: Satellite Image, ground truth, fully supervised results with our pipeline (Ours FSL), and weakly supervised results (Our segmentation). Our proposed WSL approach presents visual results comparable to more expensive FSL setups.
    }
    \label{fig:qualitative_results}
\end{figure}

\begin{figure}[t]
    \centering
    \includegraphics[width=\columnwidth]{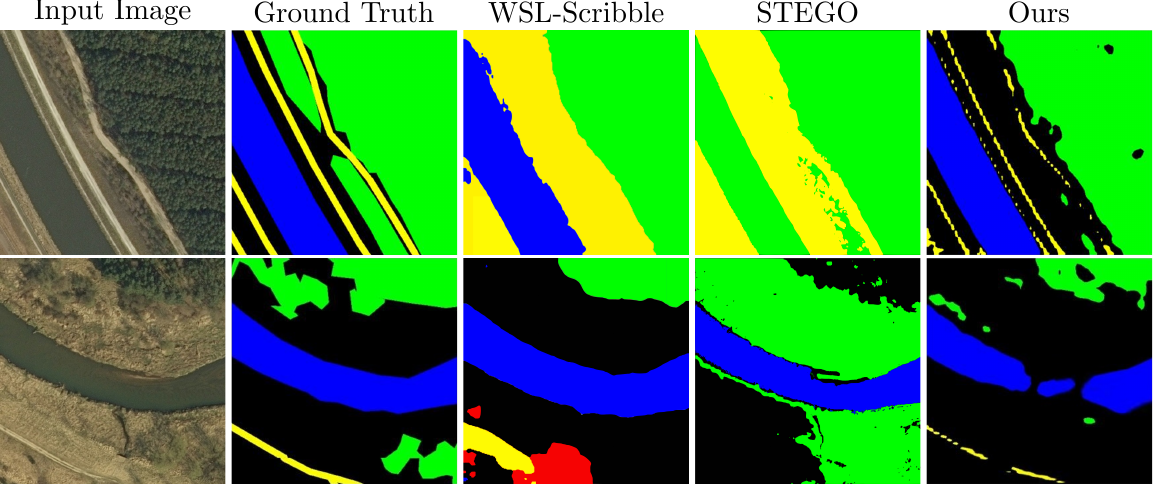}
    \caption{
    \textbf{Visual comparison with WSL methods.} We visually compare our approach with WSL-Scribble~\cite{ferreira2022weaklier} and STEGO~\cite{hamilton2022unsupervised}.
    }
    \label{fig:wsl_comparison}
    \vspace{-10pt}
\end{figure}

\mysection{Qualitative Results.}
We show some qualitative results of our approach in Fig. \ref{fig:qualitative_results} and Fig. \ref{fig:wsl_comparison}. As observed from Fig. \ref{fig:qualitative_results}, our weakly supervised semantic segmentation approach generates visually comparable results to costly, fully supervised methods. Similarly, our segmentation results shown in Fig. \ref{fig:wsl_comparison} are more consistent with the ground truth.

\mysection{Quantitative Results.}
In Table \ref{tab:comparison}, we compare our proposed approach against other Fully and Weakly supervised approaches. From the Fully supervised approaches, we compare against SGBNet, DBPNet, MAFNet~\cite{miao2023mafnet}, and our approach when using the ground truth as pseudo annotation mask (Ours-FSL), see Fig.~\ref{fig:pipeline_figure}. Similarly, we compare our approach against these weakly supervised methods: WSL-Scribble~\cite{ferreira2022weaklier}, and STEGO~\cite{hamilton2022unsupervised}. As observed, our WSL approach outperforms the other methods in the same category, and our FSL achieves competitive results with other fully supervised methods.




\vspace{-5pt}
\section{Conclusions}
\label{sec:conclusions}
In this paper, we presented a simple framework for semantic segmentation that significantly reduces the time and cost required for manual annotation. Our approach achieves high segmentation performance using only point-based annotations, minimizing annotation efforts. 
\vspace{-5pt}
\bibliographystyle{IEEEbib}
\bibliography{refs}

\end{document}